\documentclass[pdflatex,sn-mathphys-num]{sn-jnl}
\usepackage{graphicx}
\usepackage{multirow}
\usepackage{amsmath,amssymb,amsfonts}
\usepackage{amsthm}
\usepackage{mathrsfs}
\usepackage[title]{appendix}
\usepackage{xcolor}
\usepackage{textcomp}
\usepackage{manyfoot}
\usepackage{booktabs}
\usepackage{algorithm}
\usepackage{algorithmicx}
\usepackage{algpseudocode}
\usepackage{listings}
\theoremstyle{thmstyleone}

\theoremstyle{thmstyletwo}

\theoremstyle{thmstylethree}

\begin{document}

\title[Fine-Tuning Whisper for Automatic Speech Recognition in Baniwa: A Preliminary Study]{Fine-Tuning Whisper for Automatic Speech Recognition in Baniwa: A Preliminary Study}

\author*[1,2]{\fnm{Leonardo} \sur{Duart}}\email{leonardo.duart@hotmail.com}

\author[2,3]{\fnm{Tiago} \sur{Fonseca}}\email{fonsecafga@unb.br}

\author[1,2]{\fnm{Thiago} \sur{Chacón}}\email{thiagocostachacon@gmail.com}

\affil*[1]{\orgdiv{Department of Statistics}, \orgname{University of Brasilia}, \orgaddress{\city{Brasilia}, \country{Brazil}}}

\abstract{

Automatic Speech Recognition (ASR) technologies have achieved remarkable performance in recent years through the use of large multilingual foundation models. However, most advances remain concentrated on high-resource languages, while indigenous languages continue to suffer from a lack of speech resources and language technologies.

This work presents a preliminary study on the adaptation of Whisper for Automatic Speech Recognition in Baniwa, an indigenous Arawakan language spoken in Brazil, Colombia, and Venezuela. The experiments were conducted using a corpus of 1,373 manually transcribed recordings obtained from a linguistic documentation project. The corpus contains approximately 0.54 hours of speech and consists primarily of isolated words and short elicited utterances.

The Whisper Small model was fine-tuned using supervised learning and evaluated using Word Error Rate (WER) and Character Error Rate (CER). The best model achieved a WER of 37.5\% and a CER of 7.45\%, demonstrating that multilingual foundation models can be successfully adapted to extremely low-resource indigenous languages.

The results establish an initial baseline for Baniwa Automatic Speech Recognition and provide a foundation for future research involving larger datasets, language-specific adaptation strategies, and post-processing techniques.

}

\keywords{Automatic Speech Recognition, Whisper, Baniwa, Indigenous Languages, Low-Resource Languages, Speech Technology}

\maketitle

\section{Introduction}\label{sec1}

Automatic Speech Recognition (ASR) systems have experienced substantial advances in recent years, largely driven by deep learning architectures and the availability of large-scale multilingual speech corpora. Models such as Whisper have demonstrated strong performance across a wide range of languages and speech recognition tasks, establishing new benchmarks for multilingual ASR \cite{radford2023whisper}. Despite these advances, speech technology remains unevenly distributed across languages, and many indigenous and low-resource languages remain underrepresented in ASR research and have limited access to digital language resources \cite{besacier2014lr,adams2022indigenous}.

This imbalance is particularly evident in the context of indigenous languages spoken in the Amazon region. The scarcity of transcribed speech corpora, linguistic resources, and computational tools limits the development of technologies that could support language documentation, education, and language revitalization efforts. As a result, many communities remain excluded from recent advances in speech and language processing.

Baniwa is an indigenous Arawakan language spoken in the Upper Rio Negro region across Brazil, Colombia, and Venezuela. Its linguistic background, sociolinguistic context, and orthographic characteristics relevant to ASR are discussed in Section~\ref{sec:baniwa}.

Recent advances in multilingual ASR offer an opportunity to support the documentation and preservation of indigenous languages. Large pre-trained speech models can be adapted to low-resource settings through fine-tuning on relatively small datasets \cite{besacier2014lr,radford2023whisper}, reducing the amount of language-specific data required to build functional speech recognition systems. This approach is especially promising for languages such as Baniwa, where collecting large-scale speech corpora remains a challenging and resource-intensive task.

In this work, we investigate the adaptation of the Whisper model to the Baniwa language through supervised fine-tuning on a manually transcribed speech corpus. Our objective is to evaluate the feasibility of developing an ASR system for Baniwa using a limited amount of training data. We describe the corpus, the training procedure, and the resulting recognition performance measured through standard ASR evaluation metrics. This study represents an initial step toward the development of speech technologies for Baniwa and provides a foundation for future work involving larger datasets, language-specific post-processing techniques, and broader evaluations.

\section{Related Work}

\subsection{Automatic Speech Recognition for Low-Resource Languages}

Recent advances in Automatic Speech Recognition (ASR) have been largely driven by deep neural networks trained on large-scale speech corpora. While these approaches have achieved remarkable performance for high-resource languages such as English, Spanish, and Mandarin, many of the world's languages remain underrepresented in speech technology research.

Developing ASR systems for low-resource languages presents several challenges, including limited amounts of transcribed speech, scarce linguistic resources, and reduced computational infrastructure \cite{besacier2014lr}. Traditional ASR pipelines often require substantial quantities of annotated data, making them difficult to apply in indigenous and minority language contexts.

To address these limitations, transfer learning and multilingual training strategies have been widely investigated for low-resource ASR \cite{besacier2014lr}. By transferring representations learned from better-resourced languages, these approaches can reduce the amount of target-language annotated data required to develop speech recognition systems.

\subsection{Whisper and Multilingual Speech Recognition}

Whisper is a multilingual speech recognition model introduced by OpenAI and trained on approximately 680,000 hours of supervised multilingual and multitask speech data \cite{radford2023whisper}. Unlike conventional ASR systems that rely heavily on language-specific resources, Whisper leverages large-scale pre-training to achieve robust performance across multiple languages and domains.

Recent studies have investigated supervised fine-tuning of Whisper for low-resource ASR and reported substantial improvements over the corresponding pre-trained baselines \cite{liu2024whisper,ghimire2024whisper}. By leveraging representations learned during large-scale multilingual pre-training, Whisper can be adapted using comparatively limited amounts of transcribed target-language speech.

These characteristics make Whisper a promising candidate for speech recognition tasks involving indigenous and under-resourced languages, where collecting large speech corpora remains costly and time-consuming.

\subsection{Speech Technologies for Indigenous Languages}

Despite growing interest in language technologies for endangered and indigenous languages, speech resources remain scarce for many linguistic communities \cite{adams2022indigenous}. Existing work on indigenous language technologies includes the development of corpora, speech recognition systems, and language documentation tools, although technological coverage remains limited across many linguistic communities \cite{adams2022indigenous}.

In recent years, multilingual foundation models have enabled new possibilities for indigenous language processing by reducing the dependence on large language-specific datasets. Such technologies can support language documentation, educational initiatives, and the creation of digital resources that contribute to language preservation.

To the best of our knowledge, relatively few studies have investigated the application of modern ASR models to indigenous languages of the Amazon region. Consequently, evaluating the performance of multilingual speech recognition models on languages such as Baniwa remains an important research direction.

The present study contributes to this growing body of work by exploring the adaptation of Whisper to Baniwa, an indigenous Arawakan language spoken in the Upper Rio Negro region.

\section{The Baniwa Language}\label{sec:baniwa}

\subsection{Linguistic Background}

Baniwa is an indigenous language belonging to the Arawakan language family \cite{aikhenvald1999baniwa} and is spoken in the Upper Rio Negro region of the northwestern Amazon. Its speakers are distributed across Brazil, Colombia, and Venezuela \cite{aikhenvald1999baniwa}, particularly in communities located along the Içana River and its tributaries.

The language is part of a broader network of Arawakan languages that occupy large portions of South America. Despite sharing common historical origins with other Arawakan languages, Baniwa presents distinct phonological, lexical, and grammatical characteristics that contribute to its linguistic identity.

Like many indigenous languages of the Amazon, Baniwa has received increasing attention from linguists and educators due to its cultural importance and its role in preserving traditional knowledge. Efforts involving language documentation, literacy programs, and educational materials have contributed to the maintenance and transmission of the language across generations.

\subsection{Sociolinguistic Context}

Although Baniwa remains relatively vital compared to many endangered indigenous languages, important sociolinguistic changes have been observed in some communities. In particular, language shift toward Portuguese and Nheengatu has occurred in specific regions, reducing the use of Baniwa in daily communication.

At the same time, there is growing interest among indigenous communities, schools, and local organizations in strengthening the use of the native language. This demand has motivated the development of educational materials, literacy initiatives, and language documentation projects aimed at supporting language maintenance and cultural preservation.

The availability of speech technologies may contribute to these efforts by facilitating the creation of transcriptions, digital archives, educational resources, and linguistic documentation. Consequently, the development of Automatic Speech Recognition systems for Baniwa represents not only a technological challenge but also a potential contribution to language preservation initiatives.

\subsection{Orthographic Characteristics}

The orthography of Baniwa contains several characteristics that may present challenges for Automatic Speech Recognition systems.

One relevant feature is vowel length. In the Baniwa orthography used in this corpus, long vowel sounds are represented by doubling the corresponding vowel grapheme, as in \textit{aa}, \textit{ee}, \textit{ii}, \textit{oo}, and \textit{uu}. Thus, the discussion here concerns vowel sounds and their orthographic representation in Baniwa rather than a universal set of vowel letters. Correctly identifying vowel length is important because it may affect lexical distinctions.

Another characteristic is the occurrence of aspirated consonants, which are represented orthographically through the addition of the letter \textit{h}. These sounds may be acoustically subtle and therefore difficult for speech recognition systems to distinguish reliably.

The language also employs digraphs and other multi-character orthographic representations that correspond to single phonological units. Such patterns may increase the complexity of transcription tasks when compared to languages with more transparent grapheme-to-phoneme correspondences.

Finally, certain consonant distinctions are represented through doubled letters, including occurrences of double \textit{t}. These structures can be particularly challenging for ASR systems because the acoustic differences associated with consonant length or articulation may be difficult to capture, especially when training data are limited.

These orthographic and phonological characteristics make Baniwa an interesting case study for evaluating the performance of multilingual ASR models in low-resource settings.

\section{Corpus Description}\label{sec:corpus}

\subsection{Data Collection}

The speech corpus used in this study was obtained from the Baniwa-Koripako Multimedia Dictionary project, a digital language documentation initiative developed to preserve and disseminate lexical and cultural knowledge of the Baniwa language \cite{baniwadictionary}. The recordings were collected by researchers with expertise in indigenous languages and language documentation.

The corpus consists primarily of elicited speech recordings rather than spontaneous conversations. During the recording process, speakers were asked to produce specific lexical items and short expressions in Baniwa, resulting in a collection dominated by isolated words and short utterances. Consequently, the corpus differs from conversational speech datasets commonly used in Automatic Speech Recognition research.

All recordings were accompanied by manual transcriptions provided through the original dictionary documentation project \cite{baniwadictionary}. The transcriptions were produced as part of the original documentation effort and constitute the ground-truth annotations employed throughout this study.

\subsection{Corpus Statistics}

The final dataset contains 1,373 audio recordings and their corresponding transcriptions. The corpus is relatively small, totaling approximately 0.54 hours (32.4 minutes) of speech data. The recordings are predominantly composed of isolated lexical items and short utterances obtained through elicitation procedures.

Audio duration ranges from 0.48 to 5.09 seconds, with a mean duration of 1.42 seconds and a median duration of 1.37 seconds. These statistics indicate that the corpus primarily consists of short speech segments rather than continuous speech or conversational interactions.

The dataset was randomly divided into training, validation, and test subsets. Approximately 90\% of the recordings were allocated to the training set, while 5\% were reserved for validation and 5\% for testing.

Table \ref{tab:corpus} summarizes the main characteristics of the corpus.

\begin{table}[h]
\centering
\caption{Summary of the Baniwa speech corpus.}
\label{tab:corpus}
\begin{tabular}{ll}
\toprule
Characteristic & Value \\
\midrule
Language & Baniwa \\
Number of recordings & 1,373 \\
Total duration & 0.54 h \\
Mean duration & 1.42 s \\
Median duration & 1.37 s \\
Minimum duration & 0.48 s \\
Maximum duration & 5.09 s \\
Training split & 90\% \\
Validation split & 5\% \\
Test split & 5\% \\
Recording type & Elicited speech \\
Transcriptions & Manual \\
Original sampling rate & 16 kHz (796); 44.1 kHz (577) \\
Bit depth & 16 bits \\
Number of channels & Mono (796); stereo (577) \\
Audio format & PCM WAV \\
\botrule
\end{tabular}
\end{table}

\section{Methodology}

\subsection{Whisper Model}

The experiments were conducted using the Whisper Small model developed by OpenAI \cite{radford2023whisper}. Whisper is a transformer-based encoder--decoder architecture trained on approximately 680,000 hours of multilingual and multitask speech data. Its large-scale pre-training enables the model to perform speech recognition across a wide range of languages and acoustic conditions.

In this study, Whisper Small was selected as a compromise between computational efficiency and recognition performance, making it suitable for experimentation in low-resource settings.

\subsection{Preprocessing}

All audio recordings were converted to a sampling rate of 16 kHz, corresponding to the input requirements of the Whisper architecture. Speech signals were processed using the Whisper feature extractor, which converts audio waveforms into log-Mel spectrogram representations.

The corresponding transcriptions were tokenized using the Whisper tokenizer. No language-specific normalization, spelling correction, or post-processing techniques were applied. The objective was to evaluate the performance of the pre-trained model after supervised fine-tuning using only the available Baniwa speech corpus.

Since Baniwa is not officially supported among the predefined Whisper language tokens, the tokenizer was configured using Spanish transcription prompts. This strategy was adopted to provide a consistent decoding configuration during training while allowing the model to adapt to the target language through fine-tuning.

\subsection{Fine-Tuning Procedure}

Supervised fine-tuning was performed using the Hugging Face Transformers framework. The dataset was divided into training, validation, and test subsets following the proportions described in Section~\ref{sec:corpus}.

Training was conducted using a learning rate of $10^{-5}$, batch size of 8 samples, and gradient accumulation over two optimization steps. Mixed-precision training (FP16) was employed whenever supported by the available hardware.

The model was trained for 300 optimization steps. Preliminary experiments with longer training schedules indicated signs of overfitting, with no consistent improvement in validation performance. Therefore, the 300-step configuration was adopted as the final setting for this preliminary study.

Model performance was evaluated periodically during training using the validation set. No external language model, pronunciation lexicon, data augmentation strategy, or post-processing method was incorporated into the training pipeline. Consequently, the reported results correspond exclusively to the performance achieved through Whisper fine-tuning.

\section{Experimental Setup}

\subsection{Training Configuration}

Table \ref{tab:training} summarizes the main hyperparameters adopted during the fine-tuning process.

\begin{table}[h]
\centering
\caption{Training configuration used for Whisper fine-tuning.}
\label{tab:training}
\begin{tabular}{ll}
\toprule
Parameter & Value \\
\midrule
Model & Whisper Small \\
Learning rate & $1 \times 10^{-5}$ \\
Batch size & 8 \\
Gradient accumulation steps & 2 \\
Maximum training steps & 300 \\
Evaluation interval & 100 steps \\
Checkpoint interval & 100 steps \\
Mixed precision & FP16 \\
Sampling rate & 16 kHz \\
Framework & Hugging Face Transformers \\
\botrule
\end{tabular}
\end{table}

Training was performed using the Hugging Face Transformers library. Model checkpoints were evaluated and saved every 100 optimization steps. The best-performing checkpoint was selected based on validation performance.

\subsection{Evaluation Metrics}

The system was evaluated using Word Error Rate (WER) and Character Error Rate (CER), two standard metrics widely employed in Automatic Speech Recognition research.

WER measures the proportion of word-level errors between the reference transcription and the predicted transcription. It is computed as

\begin{equation}
WER = \frac{S + D + I}{N},
\end{equation}

Here, $S$ denotes substitutions, $D$ denotes deletions, $I$ denotes insertions, and $N$ denotes the total number of words in the reference transcription.

Because the corpus used in this study consists predominantly of isolated words and short utterances, Character Error Rate (CER) was also considered. CER evaluates transcription quality at the character level and is calculated as

\begin{equation}
CER = \frac{S_c + D_c + I_c}{N_c},
\end{equation}

Here, $S_c$, $D_c$, and $I_c$ denote character substitutions, deletions, and insertions, respectively, while $N_c$ denotes the total number of characters in the reference transcription.

Lower values of WER and CER indicate better recognition performance.

\section{Results}

\subsection{Training Performance}

Table \ref{tab:results} summarizes the training and validation metrics observed during the fine-tuning process.

\begin{table}[h]
\centering
\caption{Training and validation results during Whisper fine-tuning.}
\label{tab:results}
\begin{tabular}{ccccc}
\toprule
Step & Training Loss & Validation Loss & WER & CER \\
\midrule
100 & 1.0883 & 0.4313 & 0.5500 & 0.1005 \\
200 & 0.1454 & 0.3695 & 0.3750 & 0.0745 \\
300 & 0.0669 & 0.3652 & 0.4000 & 0.0728 \\
\botrule
\end{tabular}
\end{table}

The training loss decreased substantially throughout the optimization process, falling from 1.0883 at step 100 to 0.0669 at step 300. A similar trend was observed for the validation loss, which decreased from 0.4313 to 0.3652.

Recognition performance also improved considerably during training. Word Error Rate (WER) decreased from 55.0\% at step 100 to 37.5\% at step 200, while Character Error Rate (CER) decreased from 10.1\% to 7.5\%.

\subsection{Recognition Performance}

The best Word Error Rate was obtained at step 200, reaching 37.5\%. Character Error Rate continued to decrease slightly until step 300, reaching a minimum value of 7.3\%.

The results indicate that the Whisper model was able to adapt to Baniwa despite the extremely limited amount of training data available. Considering that the corpus contains only 0.54 hours of speech recordings, the observed performance demonstrates the potential of large multilingual pre-trained models for low-resource indigenous languages.

\subsection{Learning Behavior}

Although validation loss continued to decrease between steps 200 and 300, Word Error Rate increased from 37.5\% to 40.0\%. This behavior suggests that additional training did not necessarily translate into improved transcription quality.

The divergence between loss and WER may indicate the onset of overfitting, a common phenomenon when fine-tuning large neural models on small datasets. Consequently, the checkpoint obtained at step 200 was considered the most effective model according to the primary evaluation metric adopted in this study.

Overall, the results demonstrate that meaningful speech recognition performance can be achieved for Baniwa even under an extremely low-resource scenario consisting of approximately 32 minutes of manually transcribed speech.

\section{Discussion}

The results obtained in this study demonstrate that Whisper can be successfully adapted to Baniwa despite the extremely limited amount of available training data. Using only 0.54 hours of manually transcribed speech, the model achieved a minimum Word Error Rate of 37.5\% and a Character Error Rate below 8\%, indicating that meaningful speech recognition performance can be obtained even in highly constrained low-resource scenarios.

One factor that likely contributed to these results is the nature of the corpus itself. Unlike many ASR datasets composed of spontaneous speech or conversational interactions, the Baniwa corpus consists primarily of elicited lexical items and short utterances. The average recording duration is approximately 1.4 seconds, reducing linguistic and acoustic variability when compared to continuous speech recognition tasks. Consequently, the reported performance should be interpreted within the context of isolated-word and short-utterance recognition rather than large-vocabulary continuous speech recognition.

The training dynamics also suggest that the small size of the corpus imposes important limitations on model adaptation. While both training and validation losses continued to decrease throughout optimization, the best WER was achieved at step 200. Additional training resulted in lower loss values but did not improve word recognition performance, suggesting the onset of overfitting. This behavior is consistent with previous observations in low-resource speech recognition, where large neural architectures can rapidly memorize small datasets.

Several linguistic characteristics of Baniwa may also contribute to recognition errors. The language contains long vowels represented through vowel duplication, aspirated consonants, digraphs, and consonant length distinctions represented orthographically by doubled letters. These features introduce transcription challenges that may not be fully captured by a multilingual model originally trained on predominantly high-resource languages.

Despite these limitations, the results demonstrate the potential of multilingual foundation models for indigenous language technologies. The ability to obtain usable recognition performance from only a few hundred recordings highlights the value of transfer learning approaches for languages with scarce digital resources.

Future work should investigate the use of larger corpora, language-specific normalization procedures, post-processing techniques, and external language models. In addition, a detailed error analysis would help identify the linguistic phenomena that most strongly affect recognition performance in Baniwa. Such developments could contribute not only to improved ASR accuracy but also to broader efforts in language documentation, education, and digital resource creation for indigenous communities.

\section{Conclusion}

This study presented a preliminary investigation of Whisper fine-tuning for Automatic Speech Recognition in Baniwa, an indigenous Arawakan language spoken in the Upper Rio Negro region. Using a corpus composed of 1,373 manually transcribed recordings, corresponding to approximately 32 minutes of speech, we evaluated the feasibility of adapting a multilingual foundation model to an extremely low-resource language scenario.

The experimental results demonstrated that Whisper Small can successfully learn useful acoustic and linguistic representations from a relatively small amount of training data. The best-performing model achieved a Word Error Rate of 37.5\% and a Character Error Rate of 7.45\%, indicating that meaningful recognition performance can be obtained even when only limited speech resources are available.

The study also revealed limitations associated with the small size of the corpus. Although training and validation losses continued to decrease throughout optimization, recognition performance did not improve beyond 200 training steps, suggesting the onset of overfitting. These findings highlight the challenges of adapting large neural models to low-resource settings while simultaneously demonstrating the effectiveness of transfer learning approaches.

Overall, the results support the potential of multilingual ASR models for indigenous language technologies and contribute an initial benchmark for Baniwa speech recognition. Future work should explore larger speech corpora, detailed error analyses, language-specific post-processing strategies, and external language models to further improve transcription accuracy and support language documentation efforts.

\backmatter

\section*{Declarations}

\textbf{Funding}

The authors received no specific funding for this work.

\vspace{0.5em}

\textbf{Conflict of interest}

The authors declare that they have no conflict of interest.

\vspace{0.5em}

\textbf{Ethics approval}

Not applicable.

\vspace{0.5em}

\textbf{Consent to participate}

Not applicable.

\vspace{0.5em}

\textbf{Consent for publication}

Not applicable.

\vspace{0.5em}

\textbf{Data availability}

The speech corpus used in this study was obtained from an ongoing linguistic documentation project. Due to ownership and community access considerations, the data are not publicly available. Access may be granted upon reasonable request and subject to authorization from the data custodians.

\vspace{0.5em}

\textbf{Code availability}

The source code used for model training and evaluation is available from the corresponding author upon reasonable request.

\vspace{0.5em}

\textbf{Author contributions}

Leonardo Duart conducted the experiments, implemented the computational pipeline, analyzed the results, and drafted the manuscript. Tiago Fonseca supervised the machine learning and statistical aspects of the study and contributed to manuscript revision. Thiago Chacón contributed linguistic expertise, provided access to the Baniwa corpus, and revised the manuscript. All authors reviewed and approved the final version of the manuscript.

\bibliography{sn-bibliography}

\end{document}